# SciAnnotate: A Tool for Integrating Weak Labeling Sources for Sequence Labeling

**Mengyang Liu**[*]   **Haozheng Luo**[*]   **Leonard Thong**[*]   **Yinghao Li**   **Chao Zhang**   **Le Song**

```
            Georgia Institute of Technology
                 College of Computing
{mliu444,lthong3,robin1997,yinghaoli,chaozhang}@gatech.edu
                  lsong@cc.gatech.edu
```

## Abstract

Weak labeling is a popular weak supervision strategy for Named Entity Recognition (NER) tasks, with the goal of reducing the necessity for hand-crafted annotations. Although there are numerous remarkable annotation tools for NER labeling, the subject of integrating weak labeling sources is still unexplored. We introduce a web-based tool for text annotation called **SciAnnotate**, which stands for scientific annotation tool. Compared to frequently used text annotation tools, our annotation tool allows for the development of weak labels in addition to providing a manual annotation experience. Our tool provides users with multiple user-friendly interfaces for creating weak labels. SciAnnotate additionally allows users to incorporate their own language models and visualize the output of their model for evaluation. In this study, we take multi-source weak label denoising as an example, we utilized a Bertifying Conditional Hidden Markov Model to denoise the weak label generated by our tool. We also evaluate our annotation tool against the dataset provided by Mysore which contains 230 annotated materials synthesis procedures. The results shows that a 53.7% reduction in annotation time obtained AND a 1.6% increase in recall using weak label denoising. Online demo is available at https://sciannotate.azurewebsites.net/(demo account can be found in README), but we don't host a model server with it, please check the README in supplementary material for model server usage.

## 1 Introduction

Annotation is one of the most costly and labor-intensive aspects of natural language processing research, and it can be significant because the performance of language models is heavily dependent on the quality of the dataset. Despite the fact that many annotation tools (Stenetorp et al., 2012; Nakayama et al., 2018; Daudert, 2020) are designed to increase the efficiency of annotation, they nevertheless require annotators to read each word of the text attentively to preserve quality. Manual annotation at the instance level maintains the high quality of the obtained labeled data, but it is prohibitively expensive and not scalable.

Weak labeling (Lison et al., 2020) could be a more promising approach than manual annotation. It is logical that the pattern employed by annotators to annotate a word or sentence may appear several times and conform to the same set of rules.

On the basis of the preceding research, we offer SciAnnotate, a tool that simplifies the process of annotating scientific literature by providing weak labeling interfaces while preserving the superior manual annotation experience. Our system is distinguished by the following features:

- Multiple interfaces for weak labeling: SciAnnotate supports Raw Text Match, Regular Expression Match and Labeling Function.

- Integration support for Language Model: SciAnnotate allows users to deploy their language model and visualize the output on current articles.

## 2 Related Work

Annotation tools are crucial for building research datasets. There are several well-known annotation tools, including Brat (Stenetorp et al., 2012), Doccano (Nakayama et al., 2018), XConc tool (Kim et al., 2008), AWOCATo (Daudert, 2020) and MyMiner (Salgado et al., 2012). Regardless of the fact that the majority of currently available annotation tools feature high-quality labeling visualization with non-programmer-friendly user interfaces, manual annotation is still expensive and exhausting.

## 3 SciAnnotate

SciAnnotate is based on the popular annotation tool Brat, which has a substantial user base. Users who are familiar with Brat can import their data WITHOUT CHANGE by using what they already have in Brat.

---
[*] These authors contributed equally to this work

### 3.1 Weak Labeling Support

Manual annotation is quite labor-intensive and expensive for building a dataset, which can only be affordable by some big research groups. Compared to manual annotation, weak labeling could be much cheaper and efficient at the cost of weak quality guarantee. Recent research (Lison et al., 2020) also showed that weak labeling can work well with NER task. In SciAnnotate, we provide three different interfaces for users to create weak labels, (1) **Raw Text Match**, this interfaces only require domain knowledge, it will find all matched text span in the document and assign the label. (2) **Regular Expression Match**, Regular expression match requires user to be familiar with regular expression. User can annotate certain pattern with a single regular expression formula. User can also leverage existing regular expression to annotate the document. Figure. 1 and Figure. 2 is an example for detecting the percentage in document. Note that when we talk to **Text Search Match** in the following sections, we're referring to the combination of Raw Text Match and Regular Expression Match. (3) **Labeling Function**, labeling function requires require user to be familiar with Python Programming. But it can also provide more complicated pattern match, which can improve the quality of weak labels. Figure. 3 shows an naive example of labeling function which aims to detect the word *spring* and try to distinguish the meaning related to Season from the one related to Mechanical-device. The result is shown in Figure. 3. The detail of the design and format of labeling function can be found in **README** in Supplementary Materials.

### 3.2 Language Model Support

SciAnnotate offers a dedicated model server that provides services relevant to language models. Some popular annotation systems, such as Doccano and AWOCATo, incorporate third-party NLP libraries to provide comparable functionality, such as automatic sequence tagging. However, the above-mentioned tools only support certain libraries or functionalities and are completely hardcoded, making it difficult for users to alter the interface to their liking. Our model server provides an environment and a request handler at the top level, allowing users to rewrite the processing logic in the request handler to do particular tasks. Python is used to develop the model server, allowing researchers to reuse their code for data processing and evaluation.

We can just move our new models to a subfolder, package the prediction portion as a standalone API, and connect it to a new request handler. Another statistic is that it allows us to install the model server on a single server equipped with GPU to handle significant computing loads. Additionally, we may share the API with others and enable them to use our model for their own research purposes, such as comparing model performance and visualizing it.

### 3.3 Weak Label Denoising

Moreover, SciAnnotate can readily integrate multi-source label denoising models such as the hidden Markov model (HMM) (Lison et al., 2020), linked HMM (Safranchik et al., 2020) and Snorkel (Ratner et al., 2017) or distantly-supervised models such as AutoNER (Shang et al., 2018) and BOND (Liang et al., 2020). These methods take one or more labeling functions as weak supervision sources, using the results of these functions as corpus annotations to train a neural network model. Since the labeling functions, as well as their annotations, are defined by the annotators and stored in our system, we can directly modify them into the proper format, train the weakly-supervised models, and get their label predictions in our system.

## 4 Case Study

Three case studies will be discussed in this section. In the first case, the annotation speeds of Brat and SciAnnotate are compared. The second case describes a particular application of Labeling Function known as Dictionary Labeling. The last case is designed to demonstrate how language model integration and label denoising may enhance our generated weak label.

### 4.1 Annotation Support

SciAnnotate was created to improve the efficiency with which research groups from across the globe annotate documents. Annotators may annotate a document using weak labeling sources, such as text search and labeling function, with the use of the annotation tool. We conduct an experiment to compare the time required to annotate a document using our annotation tool with the conventional annota-tion tool used in industry. We have selected *The material science procedural text corpus* (Mysore et al., 2019), a corpus that poses a unique challenge to annotation tools. The chosen text is organized with attention by domain specialists in material science. It contains 230 documents, for a total of 2113 sentences, 56510 tokens, 20849 entity mentions. Then, we ask five non-experts to jointly annotate each file and calculate the average time and recall score. The results of the assessment are shown in Table. 2. Text-Search-Annotate necessitates that the annotator utilize standard text word searching to annotate the contents and Labeling-Function-Annotate demands that the files be annotated with labeling functions. Comparing the labeling function annotating approach to the Brat (manual) annotating method, the amount of time required was almost reduced in half. The recall of labeling by function decreased by 11.22% to the recall of using Brat (manual). It is evident that our annotation tool may give a significant reduction in annotation time without sacrificing precision.

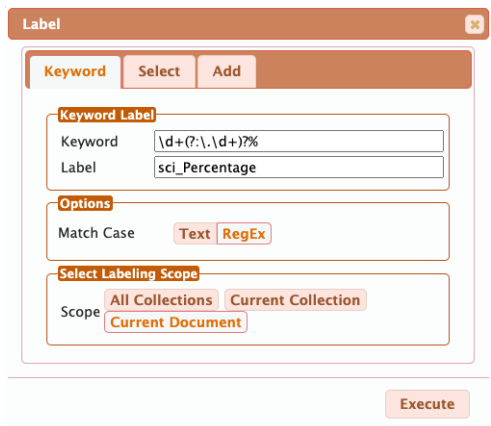

Figure 1: Regular Expression Interface

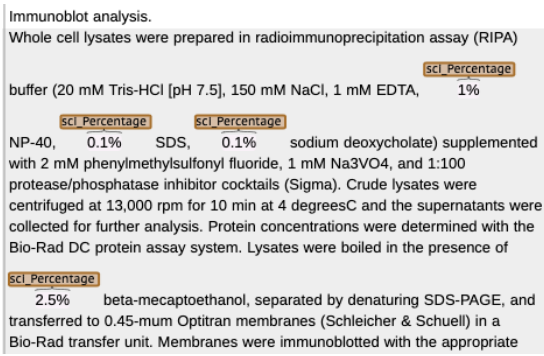

Figure 2: Regular Expression Match

| Interface  | Manual | Weak Labeling | Customized Model |
|------------|--------|---------------|------------------|
| Brat       | ✔      | ✘             | ✘                |
| Doccano    | ✔      | ✘             | ✘                |
| SciAnnotate| ✔      | ✔             | ✔                |

Table 1: Total annotation time, type selection time, and recall for different annotation methods

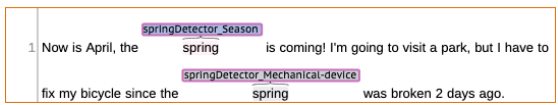

Figure 3: Labeling Function Result

### 4.2 Dictionary Labeling

Dictionary Labeling relies on the assumption that papers from the same domain will have the same annotation rules. In material science literature, for instance, *degC* will be labeled as *Condition-Unit*, and this name may appear in many files. The labeling mechanism might save a great deal of time with a pre-built lexicon. The development of such a dictionary involves specialized subject knowledge and should be carried out by domain specialists. As dictionary construction is very subjective, our tests are based on a weaker but still acceptable assumption: given a corpus from a certain domain, annotation rules from a subset of the corpus may be repeated often across the whole corpus. It implies that we might manually annotate a subset of texts before applying rules to the rest. (Figure. 4)

In the experiment, we produce a subset of the original dataset by randomly sampling the whole dataset. Then, we manually annotate the subset and construct the dictionary with the rules that exist in the subset. We use the dictionary to label the rest of the dataset. Figure. 4 depicts the recall in annotated and unannotated materials. To prevent the impact of an unbalanced number of annotations, the experiment is repeated 1,000 times for each ratio. It looks that if we annotate 15% of the whole dataset and apply the dictionary to the rest, we may extract around 47% of golden labels. It might save time for annotators. One concern is that words with numerous meanings would be incorrectly labeled; this issue might be mitigated to some degree by labeling denoising. Moreover, since manual annotation necessitates that annotators examine each word, this inaccuracy might be readily rectified.

| Method | Total(h) | Recall(%) |
|---|---|---|
| Brat | 184:21 | **92.3** |
| Doccano | 153:34 | 91.37 |
| Text-Search-Annotate | 103:37 | 84.69 |
| Labeling-Function-Annotate | 100:19 | 88.78 |
| SciAnnotate | **85:23** | 87.83 |

Table 2: Total annotation time, type selection time, and recall for different annotation methods

| Method | Recall(%) |
|---|---|
| Brat | **92.3** |
| Doccano | 91.37 |
| Labeling-Function-Annotate (Before Denoising) | 88.78 |
| Labeling-Function-Annotate (After Denoising) | 90.36 |

Table 3: The recall percentage of manual annotation, labeling-function-annotate before denoising, and labeling-function-annotate after denoising

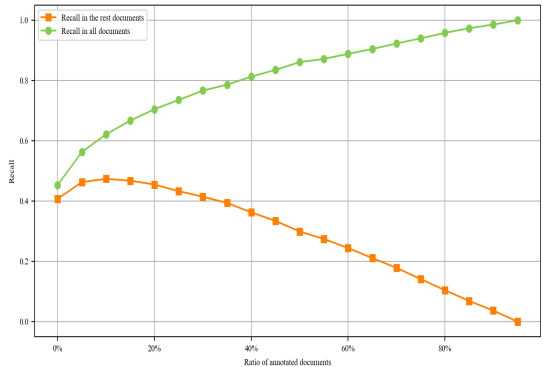

Figure 4: Recall in unannotated documents (Green); Recall in all documents (Orange)

### 4.3 Label Denoising

While efficiency is an important component of a good annotation tool, we cannot overlook the significance of annotation precision. As mentioned above, SciAnnotate allows users to integrate their own language model, thus, we took label denoising as an example. Label denoising is intended to provide the capacity to decrease the noise of weak labels, so improving the quality of our annotations. For this experiment, the same dataset as Table. 4.1 was employed, and we borrow the pre-trained model from (Li et al., 2021) to apply label denoising to weak labels. First, the text is annotated using the labeling functions. Then, we applied the denoising via model server to eliminate labeling mistakes on challenging terms such as polysemy. The assessment outcomes are shown in Table. 3 with the mean of all 230 annotation materials. The recall of the labeling function annotating technique before to denoising is 88.78% while the recall of the labeling function annotating method after denoising is 90.36%.

## 5 Conclusion and Future work

This research developed an annotation tool that integrates weak labeling sources to increase annotation efficiency with little recall loss. To take use of the strengths of various annotation methods, we provide an interface that supports a variety of labeling techniques, including manual annotation, text search annotation, and labeling function annotation. In addition, our interface can denoise weak labels using a BERT and Hidden Markov Model-based label denoising model.

Future efforts will focus on adding diverse NLP and Deep Learning technologies to enhance annotation efficiency such as suggesting labeling function. In addition, we will incorporate more Language models into our annotation tool to provide functionality such as document categorization, sentiment analysis, and translation.

## References

Tobias Daudert. 2020. A web-based collaborative annotation and consolidation tool. In *Proceedings of the 12th Language Resources and Evaluation Conference*, pages 7053–7059, Marseille, France. European Language Resources Association.

James Gardner. 2009. *The Web Server Gateway Interface (WSGI)*, pages 369–388. Apress, Berkeley, CA.

Jesse James Garrett. 2007. Ajax: A new approach to web applications.

Jin-Dong Kim, Tomoko Ohta, and Jun'ichi Tsujii. 2008. Corpus annotation for mining biomedical events from literature. *BMC bioinformatics*, 9(1):10.

Yinghao Li, Pranav Shetty, Lucas Liu, Chao Zhang, and Le Song. 2021. BERTifying the hidden Markov model for multi-source weakly supervised named entity recognition. In *Proceedings of the 59th Annual Meeting of the Association for Computational Linguistics and the 11th International Joint Conference on Natural Language Processing (Volume 1: Long Papers)*, pages 6178–6190, Online. Association for Computational Linguistics.

Chen Liang, Yue Yu, Haoming Jiang, Siawpeng Er, Ruijia Wang, Tuo Zhao, and Chao Zhang. 2020. Bond: Bert-assisted open-domain named entity recognition with distant supervision. In *Proceedings of the 26th ACM SIGKDD International Conference on Knowledge Discovery amp; Data Mining*, KDD '20, pages 1054–1064, New York, NY, USA. Association for Computing Machinery.

Pierre Lison, Jeremy Barnes, Aliaksandr Hubin, and Samia Touileb. 2020. Named entity recognition without labelled data: A weak supervision approach. In *Proceedings of the 58th Annual Meeting of the Association for Computational Linguistics*, pages 1518–1533, Online. Association for Computational Linguistics.

Sheshera Mysore, Zach Jensen, E. Kim, Kevin Huang, Haw-Shiuan Chang, Emma Strubell, J. Flanigan, A. McCallum, and Elsa Olivetti. 2019. The materials science procedural text corpus: Annotating materials synthesis procedures with shallow semantic structures. *ArXiv*, abs/1905.06939.

Hiroki Nakayama, Takahiro Kubo, Junya Kamura, Yasufumi Taniguchi, and Xu Liang. 2018. doccano: Text annotation tool for human. *Software available from https://github. com/chakkiworks/doccano*.

Alexander Ratner, Stephen H. Bach, Henry Ehrenberg, Jason Fries, Sen Wu, and Christopher Ré. 2017. Snorkel: Rapid training data creation with weak supervision. *Proc. VLDB Endow.*, 11(3):269–282.

Esteban Safranchik, Shiying Luo, and Stephen H. Bach. 2020. Weakly supervised sequence tagging from noisy rules. In *The Thirty-Fourth AAAI Conference on Artificial Intelligence, AAAI 2020, The Thirty-Second Innovative Applications of Artificial Intelligence Conference, IAAI 2020, The Tenth AAAI Symposium on Educational Advances in Artificial Intelligence, EAAI 2020, New York, NY, USA, February 7-12, 2020*, pages 5570–5578. AAAI Press.

David Salgado, Martin Krallinger, Marc Depaule, Elodie Drula, Ashish V Tendulkar, Florian Leitner, Alfonso Valencia, and Christophe Marcelle. 2012. Myminer: a web application for computer-assisted biocuration and text annotation. *Bioinformatics*, 28(17):2285–2287.

Jingbo Shang, Liyuan Liu, Xiaotao Gu, Xiang Ren, Teng Ren, and Jiawei Han. 2018. Learning named entity tagger using domain-specific dictionary. In *Proceedings of the 2018 Conference on Empirical Methods in Natural Language Processing*, pages 2054–2064, Brussels, Belgium. Association for Computational Linguistics.

Pontus Stenetorp, Sampo Pyysalo, Goran Topić, Tomoko Ohta, Sophia Ananiadou, and Jun'ichi Tsujii. 2012. brat: a web-based tool for NLP-assisted text annotation. In *Proceedings of the Demonstrations at the 13th Conference of the European Chapter of the Association for Computational Linguistics*, pages 102–107, Avignon, France. Association for Computational Linguistics.

Anne Van Kesteren. 2010. *Cross-Origin Resource Sharing*. Betascript Publishing.

## A  Implementation

Two individual servers provide the SciAnnotate service: the SciAnnotate server and the model server. The SciAnnotate server offers annotation capabilities, while the model server provides Language Model capabilities such as label denoising and model training.

The backend of SciAnnotate is written in Python with a client-server architecture. It employs the JavaScript Object Notation (JSON) data transfer standard via the HTTP protocol. The frontend, on the other hand, is constructed utilizing HTML, CSS, and JavaScript, which are relatively familiar among developers, enabling users to quickly change it. Asynchronous JavaScript and XML (AJAX) (Garrett, 2007) is used to manage communication between the frontend and backend, allowing our system to support asynchronous messaging. Our annotation tool also supports Common Gateway Interface and Fast Common Gateway Interface; using the latter may greatly cut response time and enhance the user experience for annotators. SciAnnotate additionally supports dynamic compilation, which allows annotators to design their own labeling functions and criteria. It enables real-time debugging and execution of labeling routines.

The model server that supports the SciAnnotate annotation tool is a Restful Application Programming Interface (API) written with Python using the Flask framework, which supports Web Server Gateway Interface (WSGI) (Gardner, 2009). Cross-origin resource sharing (CORS) (Van Kesteren, 2010) is necessary for the SciAnnotate server to access cross-origin model server API, since the model server often has a different address. Therefore, the model server requires a preset request data format. The SciAnnotate server will address this problem by automatically transmitting transformed data that conforms to the established format to the client, which will then forward it to the model server. The model server will then deliver formatted data that will be rendered directly by the client's frontend rendering component. We published our Docker image for deployment purposes. It incorporates the most widely used Deep Learning components and libraries, allowing annotators to easily experiment with various models and personalize their Deep Learning features.